\documentclass[sigconf,nonacm]{acmart}

\renewcommand\footnotetextcopyrightpermission[1]{}
\renewcommand{\shortauthors}{}

\usepackage{amsmath}

\usepackage{amssymb}
\usepackage{algorithm}
\usepackage{algorithmic}
\usepackage{booktabs}
\usepackage{multirow}
\usepackage{graphicx}
\usepackage{float}

\newcommand{\RQ}[1]{\textbf{RQ#1}}

\begin{document}

\begin{titlepage}
\centering
\vspace*{3cm}
{\LARGE\bfseries Estimating Visual Attribute Effects in Advertising from Observational Data: A Deepfake-Informed Double Machine Learning Approach\par}
\vspace{2.5cm}
{\large\bfseries Yizhi Liu\par}
\vspace{0.3cm}
Robert H. Smith School of Business\\
University of Maryland, College Park, MD 20742\\
\href{mailto:yizhiliu@umd.edu}{\color{blue}yizhiliu@umd.edu}
\vspace{1.5cm}

{\large\bfseries Balaji Padmanabhan\par}
\vspace{0.3cm}
Robert H. Smith School of Business\\
University of Maryland, College Park, MD 20742\\
\href{mailto:bpadmana@umd.edu}{\color{blue}bpadmana@umd.edu}
\vspace{1.5cm}

{\large\bfseries Siva Viswanathan\par}
\vspace{0.3cm}
Robert H. Smith School of Business\\
University of Maryland, College Park, MD 20742\\
\href{mailto:sviswan1@umd.edu}{\color{blue}sviswan1@umd.edu}
\vspace{2cm}

{\large\bfseries Feb 23, 2026\par}
\end{titlepage}

\title[Estimating Visual Attribute Effects in Advertising]{Estimating Visual Attribute Effects in Advertising from Observational Data: A Deepfake-Informed Double Machine Learning Approach}

\author{Yizhi Liu}
\affiliation{%
  \institution{Robert H. Smith School of Business, University of Maryland}
  \city{College Park}
  \state{MD}
  \postcode{20742}
  \country{USA}
}
\email{yizhiliu@umd.edu}

\author{Balaji Padmanabhan}
\affiliation{%
  \institution{Robert H. Smith School of Business, University of Maryland}
  \city{College Park}
  \state{MD}
  \postcode{20742}
  \country{USA}
}
\email{bpadmana@umd.edu}

\author{Siva Viswanathan}
\affiliation{%
  \institution{Robert H. Smith School of Business, University of Maryland}
  \city{College Park}
  \state{MD}
  \postcode{20742}
  \country{USA}
}
\email{sviswan1@umd.edu}

\begin{abstract}
Digital advertising increasingly relies on visual content, yet marketers lack rigorous methods for understanding how specific visual attributes causally affect consumer engagement. This paper addresses a fundamental methodological challenge: estimating causal effects when the treatment, such as a model's skin tone, is an attribute embedded within the image itself. Standard approaches like Double Machine Learning (DML) fail in this setting because vision encoders entangle treatment information with confounding variables, producing severely biased estimates. We develop DICE-DML (Deepfake-Informed Control Encoder for Double Machine Learning), a framework that leverages generative AI to disentangle treatment from confounders. The approach combines three mechanisms: (1) deepfake-generated image pairs that isolate treatment variation; (2) DICE-Diff adversarial learning on paired difference vectors, where background signals cancel to reveal pure treatment fingerprints; and (3) orthogonal projection that geometrically removes treatment-axis components. In simulations with known ground truth, DICE-DML reduces root mean squared error by 73--97\% compared to standard DML, with the strongest improvement (97.5\%) at the null effect point, demonstrating robust Type I error control. Applying DICE-DML to 232,089 Instagram influencer posts, we estimate the causal effect of skin tone on engagement. Standard DML produces diagnostically invalid results (negative outcome $R^2$), while DICE-DML achieves valid confounding control ($R^2 = 0.63$) and estimates a marginally significant negative effect of darker skin tone ($-522$ likes; $p = 0.062$), substantially smaller than the biased standard estimate. Our framework provides a principled approach for causal inference with visual data when treatments and confounders coexist within images.
\end{abstract}


\maketitle

\section{Introduction}

Visual content has become central to digital marketing. On platforms such as Instagram, Pinterest, and TikTok, images and videos drive consumer attention, engagement, and purchase decisions~\cite{li2020picture,zhang2022airbnb}. Firms invest heavily in visual creatives, and marketing analytics increasingly relies on computer vision to understand what makes visual content effective~\cite{liu2020youtube,hartmann2023sentiment}. Deep learning models can now predict engagement from images with remarkable accuracy, enabling data-driven optimization of visual marketing strategies.

Yet prediction is not causation. Modern deep learning excels at forecasting consumer response but provides limited insight into \emph{why} certain visual content performs better than others. When a marketing manager asks whether a specific visual attribute, such as a model's appearance, a color scheme, or a composition choice, causes changes in consumer engagement, predictive models cannot answer. They conflate the attribute's effect with everything else that varies alongside it. This distinction matters profoundly for managerial decision-making: acting on correlational associations could be ineffective or counterproductive if the underlying relationships are confounded rather than causal~\cite{pearl2009causality,mullainathan2017ml}.

The challenge is particularly acute when studying how attributes of people depicted in images affect consumer response. Questions about diversity and representation in advertising---whether featuring models of different skin tones, body types, or ages affects engagement---are of growing importance~\cite{appiah2001ethnic,peck2004body}. These questions have gained urgency as firms face increasing pressure to embrace inclusive representation. Yet these questions are fundamentally causal, and observational data present severe confounding challenges. Images featuring models with different characteristics also differ systematically in lighting, styling, composition, and context. This motivates our first research question (\RQ{1}): \emph{How can we estimate the causal effect of a visual attribute in advertising images from observational data?}

Double Machine Learning (DML) offers a principled framework that appears well suited to this task~\cite{chernozhukov2018dml,shi2025dml}. By combining flexible machine learning for nuisance estimation with orthogonalization and cross-fitting, DML achieves valid inference even when confounders are numerous and complex. The framework has attracted considerable attention for its ability to handle modern high-dimensional settings while maintaining inferential guarantees. Recent extensions apply DML to unstructured data by using neural network encoders to extract confounder representations from text and images~\cite{veitch2020text,klaassen2024doublemldeep}. In principle, a vision encoder could capture the complex visual factors that confound the relationship between a model's appearance and audience engagement.

However, we identify a fundamental problem with applying this approach when the treatment is an attribute within the image. Standard vision encoders (e.g., ResNet, CLIP, ViT) are trained to capture all visually salient information, including the treatment attribute itself, in a single representation. When this representation serves as a control variable, information about the treatment leaks into the controls, distorting causal estimation. We term this phenomenon \emph{visual treatment leakage}. The consequences are severe: in our empirical application, DML with standard encoders produces a negative outcome $R^2$, meaning the nuisance model predicts engagement worse than simply using the sample mean. This counterintuitive result arises because the model learns to predict treatment-induced variation rather than genuine confounding variation. This motivates our second research question (\RQ{2}): \emph{How can we construct image representations that enable valid causal inference when the treatment is embedded within the image?}

We propose DICE-DML (Deepfake-Informed Control Encoder for Double Machine Learning), a framework that addresses visual treatment leakage by leveraging generative AI to guide representation learning. Our key insight is that deepfake technology, which can modify specific visual attributes while preserving others, provides structured supervision for learning treatment-invariant representations. By generating paired images that differ primarily in the treatment attribute, we create training signals that teach the encoder to distinguish treatment from confounding information. Importantly, we do not treat these generated images as true counterfactuals; they serve only as weak supervision for representation learning, while causal identification still relies on standard conditional independence assumptions.

We evaluate DICE-DML through comprehensive simulation studies and an empirical application estimating the effect of influencer skin tone on Instagram engagement. In simulations with known ground-truth effects, DICE-DML reduces root mean squared error by 73--97\% compared to standard DML. In the empirical application, DICE-DML transforms the diagnostic profile from invalid (outcome $R^2 = -0.003$) to valid ($R^2 = 0.626$). We estimate a marginally significant negative effect of darker skin tone ($-522$ likes; $p = 0.062$), substantially smaller in magnitude than the biased estimate from standard DML ($-1,455$ likes). Our results show that naively applying powerful representation learning to causal estimation can be worse than using no controls at all, and that generative models can be repurposed to restore causal validity.

This paper makes three contributions. First, we identify visual treatment leakage as a fundamental challenge for causal inference when treatments are embedded within images, provide a formal definition, and establish diagnostic criteria for detecting this issue. Second, we develop DICE-DML, a methodological framework combining deepfake weak supervision, difference-based adversarial learning, and orthogonal projection to construct representations suitable for causal estimation. Third, we provide empirical evidence on the causal effect of skin tone representation in social media advertising, a question of substantive importance for inclusive marketing practice.

\section{Related Literature}

\subsection{Causal Inference with High-Dimensional Data}

Double Machine Learning~\cite{chernozhukov2018dml} provides a framework for causal inference when control variables are high-dimensional. The key innovation is orthogonalization: by residualizing both outcome and treatment with respect to controls, DML ensures that estimation errors in nuisance functions do not bias the treatment effect estimate. This property, known as Neyman orthogonality, allows researchers to use flexible machine learning methods while maintaining inferential validity. Combined with cross-fitting to prevent overfitting, DML achieves root-n consistent inference even in high-dimensional settings. The framework has been extended to heterogeneous effects~\cite{athey2019grf,semenova2021debiased}, panel settings~\cite{arkhangelsky2021sdid}, and dynamic treatments~\cite{lewis2021dynamic}.

Recent work extends DML to unstructured data. Veitch et al.~\cite{veitch2020text} and Keith et al.~\cite{keith2020text} develop methods for causal inference with text, using document embeddings as confounding controls. Egami et al.~\cite{egami2022text} provide a comprehensive framework for text-based causal inference. Klaassen et al.~\cite{klaassen2024doublemldeep} propose DoubleMLDeep, using neural encoders to extract representations from images and text. These approaches share a common strategy: embed unstructured data into vector representations using pretrained encoders, then apply DML to the resulting features. This works well when unstructured data contains confounders for an \emph{external} treatment. However, when the treatment is an attribute \emph{within} the unstructured data, standard encoders embed both treatment and confounders together. Our study fills this gap.

\subsection{Representation Learning for Causal Estimation}

Balanced representation learning~\cite{johansson2016learning,shalit2017estimating} reduces distributional differences between treatment groups to minimize selection bias in counterfactual prediction. The intuition is that if treated and control units appear similar in representation space, predicting counterfactual outcomes becomes more reliable. Louizos et al.~\cite{louizos2017causal} use variational autoencoders to infer latent confounders. However, the objective of balanced representation learning, distributional similarity across treatment groups, differs from DML's orthogonality requirement. Balanced representations may still encode treatment information; they equalize distributions without necessarily eliminating predictability. DML requires controls that are uninformative about treatment assignment.

Fair representation learning~\cite{zemel2013fair,edwards2016censoring,madras2018fair} pursues a related goal: learning representations that exclude sensitive attributes to prevent discrimination. While both fair representation learning and DML-based causal estimation seek attribute invariance, they impose different requirements. Fair learning constrains \emph{predictions} to be independent of the sensitive attribute; the representation may retain attribute information if it does not affect outcomes. DML estimation is more sensitive to attribute leakage: treatment information in the representation can be exploited by the propensity model, potentially biasing effect estimates.

\subsection{Visual Content Effects in Marketing}

Prior research has extensively studied how visual elements affect consumer behavior, documenting effects of color on brand perception~\cite{labrecque2012color}, visual complexity on advertising~\cite{pieters2010stopping}, human presence on social media engagement~\cite{bakhshi2014faces}, and aesthetic quality on product evaluation~\cite{hagtvedt2008art}. Deep learning has enabled extraction of visual features at scale~\cite{liu2020youtube,zhang2022airbnb,hartmann2023sentiment}, supporting large-sample studies of visual marketing effectiveness.

A growing stream examines diversity and representation in advertising~\cite{appiah2001ethnic,forehand2001priming,peck2004body,kozar2008older}. This work informs inclusive marketing strategy but faces methodological limitations. Laboratory experiments offer internal validity through random assignment but limited external validity due to artificial settings. Studies using observational data offer naturalism and scale but suffer from confounding. Our study bridges this gap, enabling causal inference from observational field data with rigorous confounding control.

\section{Methodology}

We first introduce notation and review Double Machine Learning (Section~3.1). We then formalize visual treatment leakage (Section~3.2) and present our DICE-DML framework (Section~3.3).

\subsection{Double Machine Learning}

Consider a dataset of advertising images where we wish to estimate the causal effect of a visual attribute on consumer engagement. Table~\ref{tab:notation} summarizes key notation used throughout this paper.

\begin{table}[t]
\caption{Summary of Key Notation}
\label{tab:notation}
\begin{tabular}{ll}
\toprule
\textbf{Symbol} & \textbf{Description} \\
\midrule
$T \in \{0,1\}$ & Binary treatment (e.g., skin tone) \\
$Y \in \mathbb{R}$ & Outcome (e.g., number of likes) \\
$I$ & Original image \\
$I'$ & Deepfake counterpart of $I$ \\
$Z = f(I)$ & Learned representation of image $I$ \\
$X^*$ & True (unobserved) confounders \\
$g(X) = E[Y|X]$ & Outcome nuisance function \\
$m(X) = E[T|X]$ & Propensity score function \\
$\tau$ & Average Treatment Effect (ATE) \\
$\delta$ & Treatment axis in representation space \\
$\alpha$ & Projection strength parameter \\
\bottomrule
\end{tabular}
\end{table}

Under the potential outcomes framework~\cite{rubin1974estimating}, the Average Treatment Effect (ATE), $\tau = E[Y(1)-Y(0)]$, is identified under the conditional independence assumption $(Y(1), Y(0)) \perp T \mid X$ and the overlap condition $0 < P(T = 1|X) < 1$.

DML~\cite{chernozhukov2018dml} considers the partially linear model where the outcome depends linearly on treatment but flexibly on controls. Let $g(X) = E[Y|X]$ denote the outcome nuisance function and $m(X) = E[T|X]$ denote the propensity score. The DML estimator constructs residualized outcome $\tilde{Y} = Y - \hat{g}(X)$ and residualized treatment $\tilde{T} = T - \hat{m}(X)$, then estimates:
\begin{equation}
\hat{\tau} = \left[\sum_{i=1}^{N}\tilde{T}_i^2\right]^{-1}\sum_{i=1}^{N}\tilde{T}_i\tilde{Y}_i
\label{eq:dml}
\end{equation}

This orthogonalization yields Neyman orthogonality: first-order errors in estimating the nuisance functions do not bias the treatment effect estimate. For valid estimation, the control variables $X$ must capture confounding variation but \emph{not} include components of the treatment itself. Violating this requirement induces ``bad control'' bias~\cite{angrist2009mostly}. In causal graph terms, standard representation learning encodes $T$ into $Z$, making $Z$ a descendant of $T$. Controlling for such a descendant over-controls for the treatment effect and introduces collider bias.

\subsection{Visual Treatment Leakage}

Applying DML to image data requires extracting a representation $Z = f(I)$ from image $I$ via an encoder $f$ (e.g., ResNet, CLIP). This representation then serves as the control variable $X$ in the DML framework. However, when the treatment is an attribute within the image, standard vision encoders violate DML's key requirement.

\begin{definition}[Visual Treatment Leakage]
Visual treatment leakage occurs when the learned representation $Z = f(I)$ contains information about the treatment $T$ beyond what is attributable to true confounders. Formally, leakage is present when $I(T; Z|X^*) > 0$, where $X^*$ denotes the true confounding factors and $I(\cdot)$ denotes conditional mutual information.
\end{definition}

Standard vision encoders inevitably exhibit leakage when the treatment is a visual attribute. These encoders are trained to capture all visually salient information for downstream prediction tasks. For instance, skin tone, as a visually salient attribute, is naturally encoded alongside lighting, composition, and other factors. There is no mechanism in standard encoder training to separate treatment-related visual features from confounder-related visual features.

When the representation $Z$ contains treatment information, two pathologies arise. First, the propensity model $\hat{m}(Z)$ can predict treatment with artificially high accuracy by exploiting leaked treatment signal, reducing variance of the residualized treatment $\tilde{T}$ and inflating estimator variance. Second, the outcome nuisance model $\hat{g}(Z)$ learns to predict outcome variation that is causally downstream of treatment rather than variation from true confounders, inducing bias.

A critical diagnostic for detecting visual treatment leakage is the outcome $R^2$ on held-out data. When leakage is severe, the outcome nuisance model fits treatment-induced variation rather than genuine confounding variation. This can produce \emph{negative} $R^2$: the model predicts worse than simply using the sample mean. Negative outcome $R^2$ is a red flag signaling that the representation is unsuitable for causal inference.

\subsection{DICE-DML Framework}

\begin{figure*}[t]
\centering
\includegraphics[width=0.85\textwidth]{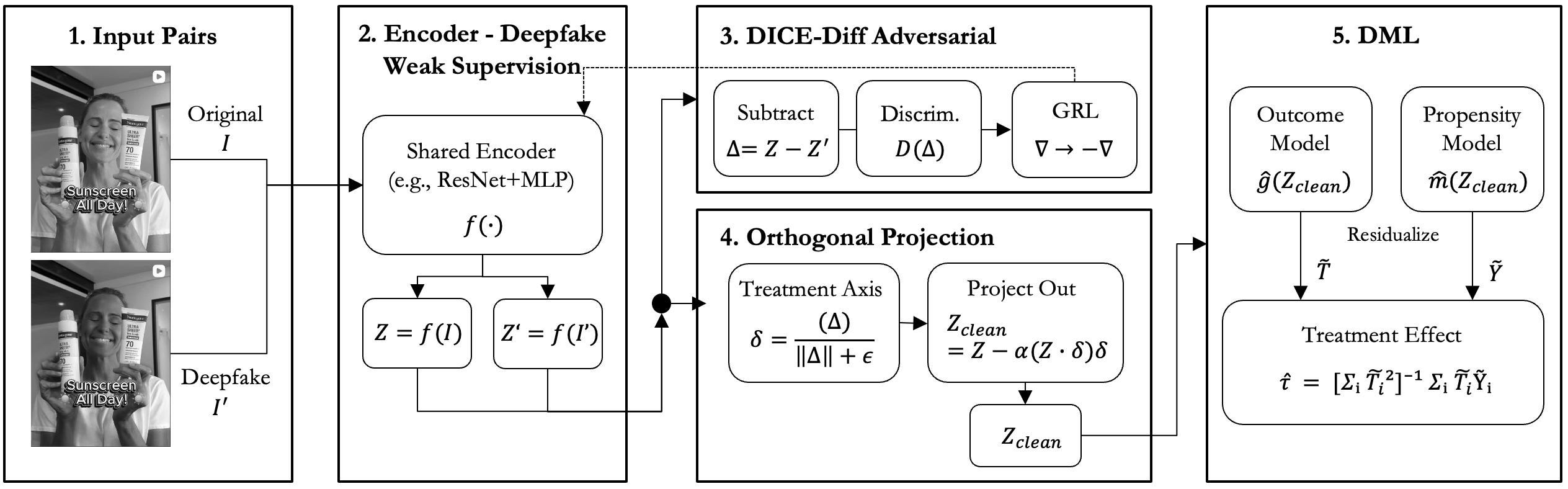}
\caption{The DICE-DML Framework. The pipeline consists of five stages: (1) Input pairs of original images and their deepfake counterparts with altered treatment attribute; (2) A shared encoder (e.g., ResNet+MLP) extracts representations $Z = f(I)$ and $Z' = f(I')$; (3) The DICE-Diff adversarial learning operates on difference vectors $\Delta = Z - Z'$ where background signals cancel, training discriminators to detect treatment while the encoder learns to fool them via gradient reversal; (4) Orthogonal projection computes sample-wise treatment directions from paired differences and projects them out to obtain $Z_{\text{clean}}$; (5) The cleaned representations feed into standard DML with outcome and propensity models for treatment effect estimation.}
\Description{A diagram showing the DICE-DML framework with five stages.}
\label{fig:framework}
\end{figure*}

We propose DICE-DML to address visual treatment leakage, as illustrated in Figure~\ref{fig:framework}. The framework operates in two stages: (1) training a specialized encoder using adversarial learning to produce representations that exclude treatment information while preserving confounding content; (2) applying orthogonal projection during both training and inference to geometrically remove treatment-axis components.

\textbf{Deepfake Weak Supervision.} The core challenge in learning treatment-invariant representations is identifying which visual features constitute treatment versus confounders. Our key insight is that deepfake technology, which can modify specific visual attributes while preserving others, provides exactly the structured supervision needed. For each original image $I$ with treatment value $T = t$, we generate a synthetic counterpart $I'$ with altered treatment $T' = 1 - t$ while preserving other visual characteristics. The resulting pair $(I, I')$ shares confounding factors but differs in treatment. This pairing provides a reference frame: features that change across the pair are treatment-related and should be excluded; features that remain constant are confounder-related and should be preserved.

We emphasize that these generated pairs provide weak supervision for representation learning, not experimental counterfactuals. Because deepfakes may introduce artifacts, we do not use them for outcome imputation; identification still relies on conditional independence given confounders. 

\textbf{DICE-Diff Adversarial Learning.} Standard adversarial training for invariance, where a discriminator tries to predict $T$ from $Z$ and the encoder tries to fool it, is contaminated by spurious correlations. If certain backgrounds systematically co-occur with certain treatment values, the discriminator can exploit these correlations rather than treatment signal itself.

We address this by having the discriminator operate on \emph{difference vectors} between paired representations. Let $Z = f(I)$ and $Z' = f(I')$ denote encoder outputs for an original image and its deepfake counterpart. Because the pair shares background characteristics, these features cancel in the difference:
\begin{equation}
Z - Z' \approx (T_{\text{features}}) - (T'_{\text{features}})
\label{eq:diff}
\end{equation}

The difference vector isolates pure treatment variation. A discriminator operating on $Z - Z'$ can only succeed by detecting treatment fingerprints; it cannot exploit background shortcuts. Through gradient reversal~\cite{ganin2016domain}, the encoder is trained to make this difference vector uninformative about treatment identity.

\textbf{Orthogonal Projection.} Beyond adversarial training, we impose a geometric constraint that directly removes treatment information. For each sample pair, we compute the difference vector and project out its component from both representations:
\begin{equation}
\Delta_i = Z_i - Z'_i, \quad \hat{\delta}_i = \Delta_i / (\|\Delta_i\| + \epsilon)
\label{eq:delta}
\end{equation}
\begin{equation}
Z_{\text{clean},i} = Z_i - \alpha(Z_i \cdot \hat{\delta}_i)\hat{\delta}_i
\label{eq:proj}
\end{equation}

where $\alpha \in [0,1]$ controls projection strength. We use $\alpha = 0.95$ to balance aggressive treatment removal against the risk of over-correction. This projection is applied during both training (to compute losses on cleaned representations) and inference (to produce final embeddings for DML).

\textbf{Training Objective.} The DICE encoder minimizes a composite loss:
\begin{equation}
\mathcal{L} = \lambda_{\text{adv}}\mathcal{L}_{\text{adv}} + \lambda_{\text{cons}}\mathcal{L}_{\text{cons}} + \lambda_{\text{var}}\mathcal{L}_{\text{var}} + \lambda_{Y}\mathcal{L}_{Y} + \lambda_{\text{ctr}}\mathcal{L}_{\text{ctr}}
\label{eq:loss}
\end{equation}

The adversarial loss $\mathcal{L}_{\text{adv}}$ implements DICE-Diff learning through two complementary discriminators. The primary discriminator $D$ is a multilayer perceptron that receives difference vectors and predicts treatment direction. We also employ a linear discriminator $D_{\text{lin}}$ consisting of a single linear layer, specifically designed to eliminate linear treatment fingerprints that logistic regression might exploit. Both discriminators are trained with binary cross-entropy loss, and through gradient reversal layers~\cite{ganin2016domain}, the encoder maximizes this loss while discriminators minimize it, creating an adversarial game that drives treatment information out of the representation.

The consistency loss $\mathcal{L}_{\text{cons}} = \frac{1}{N}\sum_{i=1}^{N}\|Z_{\text{clean},i} - Z'_{\text{clean},i}\|^2$ encourages the cleaned representations of paired images to be similar, directly penalizing residual treatment-related differences. Variance regularization $\mathcal{L}_{\text{var}}$ prevents representation collapse following VICReg~\cite{bardes2022vicreg}, penalizing low variance in each representation dimension. The outcome prediction loss $\mathcal{L}_{Y}$ ensures the representation retains information predictive of $Y$, preserving confounding information necessary for DML. The contrastive loss $\mathcal{L}_{\text{ctr}}$ requires different samples to remain distinguishable in representation space, preventing the encoder from mapping all inputs to similar outputs.

\section{Empirical Setting}

\subsection{Data}

We evaluate DICE-DML using data from Instagram influencers, estimating the causal effect of skin tone on user engagement. We use data from Kim et al.~\cite{kim2020influencer}, who collected Instagram posts across multiple influencer categories for research on influencer marketing. To focus on skin tone effects in a context where facial appearance is central, we select the Beauty category, where images prominently and consistently feature human faces and skin tone is visually salient and commercially relevant.

We apply RetinaFace~\cite{deng2020retinaface}, a state-of-the-art deep learning model for face detection, to identify images containing faces. We use a high confidence threshold of 0.9 to ensure accurate detection and avoid false positives. Images without detected faces are excluded from analysis, as skin tone cannot be measured without visible facial features. For images containing multiple faces, we select the highest-confidence detection as the focal face. After filtering, our final sample comprises 232,089 images with associated engagement metrics and influencer characteristics.

We measure skin tone using the Individual Typology Angle (ITA\textdegree), a well-established metric in dermatological research that quantifies skin lightness on a continuous scale~\cite{chardon1991skin}. ITA\textdegree{} is computed from the CIELAB color space representation of facial skin:
\begin{equation}
\text{ITA\textdegree} = \arctan[(L^* - 50) / b^*] \times (180/\pi)
\label{eq:ita}
\end{equation}

where $L^*$ is the lightness component (0 = black, 100 = white) and $b^*$ is the yellow-blue channel. Higher ITA\textdegree{} values indicate lighter skin tones. Following the classification scheme in Chardon et al.~\cite{chardon1991skin}, we define a binary treatment variable at the threshold ITA\textdegree{} $= 28$, which corresponds to the boundary between ``Intermediate'' and ``Tan'' categories. In our sample, 19.0\% of images have $T = 1$ (darker skin tone). The outcome variable $Y$ is the number of likes received by each post, with a sample mean of 6,192 and standard deviation 33,129.

\subsection{Deepfake Generation}

For each original image, we alter skin tone using LAB color space manipulation. The procedure involves four steps. First, we identify the skin region based on the face bounding box detected by RetinaFace, extended to include visible neck and shoulder areas. Second, we adjust the $L^*$ (lightness) channel to shift the measured ITA\textdegree{} across the 28\textdegree{} threshold in the opposite direction. Third, we fine-tune the $a^*$ (red-green) and $b^*$ (yellow-blue) channels to maintain natural skin appearance and avoid unnatural color casts. Fourth, we apply Gaussian feathering (kernel size 31, $\sigma = 15$) at region boundaries to ensure smooth transitions between modified and unmodified areas.

The transformation strength is calibrated adaptively to ensure the generated image's ITA\textdegree{} crosses the 28\textdegree{} threshold. If the initial transformation is insufficient, we iteratively increase intensity (scaling factors 1.0 $\rightarrow$ 1.5 $\rightarrow$ 2.0 $\rightarrow$ 2.5 $\rightarrow$ 3.0) until the threshold is crossed. This ensures that each original image is paired with a deepfake that lies on the opposite side of the treatment boundary. We note that our implementation uses LAB color space manipulation rather than neural network-based deepfake generation, primarily for computational efficiency given our dataset of over 230,000 images. This simpler approach suffices because the manipulation targets a well-defined visual attribute (skin tone lightness) and the resulting pairs need only provide approximate weak supervision for representation learning, not pixel-perfect counterfactuals. More sophisticated generative methods (e.g., diffusion models) could be employed for treatments requiring more complex manipulations.

\subsection{Implementation}

We extract 2,048-dimensional features from both original and deepfake images using ResNet-50 pretrained on ImageNet, a standard backbone for image representation. These features serve as input to the DICE encoder, which is a three-layer multilayer perceptron with architecture 2048 $\rightarrow$ 1024 $\rightarrow$ 512, batch normalization, ReLU activations, and dropout (rate 0.1) for regularization. The primary discriminator $D$ is also a multilayer perceptron with hidden layers 256 $\rightarrow$ 128 using ReLU activations, while the linear discriminator $D_{\text{lin}}$ consists of a single linear layer.

Training proceeds for 30 epochs using the Adam optimizer with a learning rate of $10^{-4}$ and a cosine annealing schedule. We use batch size 256, projection strength $\alpha = 0.95$, and loss weights $\lambda_{Y} = 10$, $\lambda_{\text{adv}} = 5$, $\lambda_{\text{cons}} = 0.5$, $\lambda_{\text{var}} = 5$, $\lambda_{\text{ctr}} = 3$. Gradient reversal uses maximum reversal strength $\alpha_{\text{GRL}} = 1.0$ with a linear warmup schedule over 3 epochs.

For the DML estimation stage, we use 5-fold cross-fitting: data are split into five folds, and nuisance functions are estimated on four folds then used to compute residuals on the held-out fold, rotating through all combinations. Unless otherwise noted, we use Random Forest with 50 trees and a maximum depth of 8 for both outcome and propensity nuisance models. We report heteroskedasticity-robust standard errors and construct 95\% confidence intervals using the normal approximation, which is valid given DML's asymptotic normality.

\section{Results}

\subsection{Simulation Validation}

We first validate DICE-DML using simulations where the ground-truth treatment effect $\tau$ is known by construction. Following the evaluation protocol of Chernozhukov et al.~\cite{chernozhukov2018dml}, we use real image features from our Instagram dataset but generate synthetic outcomes according to a data-generating process with known parameters. This approach allows us to assess estimation accuracy under realistic feature distributions while maintaining experimental control over the true effect size. We generate outcomes according to the partially linear model:
\begin{equation}
Y = \tau \cdot T + g(Z) + \varepsilon
\label{eq:dgp}
\end{equation}

where $g(Z)$ is a nonlinear function of image features capturing confounding, constructed as a combination of nonlinear transformations (hyperbolic tangent, sine, polynomial, ReLU) of random projections of $Z$ plus interaction terms. The error $\varepsilon$ is drawn from a standard normal distribution with a signal-to-noise ratio of 1.0. We inject treatment leakage into the original ResNet features (leakage strength 0.5) to simulate the real-world scenario where encoders capture treatment information.

We evaluate seven effect sizes in standardized units (Cohen's $d$): $\tau \in \{-0.30, -0.20, -0.10, 0.00, 0.10, 0.20, 0.30\}$. This range includes negative effects, positive effects, and importantly, the null effect ($\tau = 0$), which is critical for assessing Type I error control. For each value of $\tau$, we run 20 independent simulations with sample size $N = 50{,}000$, computing bias and RMSE for each method. Table~\ref{tab:simulation} presents the simulation results.

\begin{table}[t]
\caption{Simulation Results with Known Ground-Truth Effects}
\label{tab:simulation}
\begin{tabular}{lccc}
\toprule
\textbf{Method} & \textbf{Avg $|\text{Bias}|$} & \textbf{Avg RMSE} & \textbf{Reduction} \\
\midrule
Naive OLS & 0.067 & 0.168 & --- \\
DML (Original) & 0.115 & 0.317 & (baseline) \\
\textbf{DML (DICE)} & \textbf{0.051} & \textbf{0.052} & \textbf{73--97\%} \\
\bottomrule
\end{tabular}
\begin{flushleft}
\footnotesize \textit{Notes:} Average metrics across seven $\tau$ values. 20 simulations per $\tau$; $N = 50{,}000$ per simulation. RMSE reduction relative to DML (Original).
\end{flushleft}
\end{table}

Three key findings emerge from this analysis. First, DICE-DML dramatically reduces estimation error. Average RMSE falls from 0.317 for standard DML to 0.052 for DICE-DML, representing an 84\% reduction in root mean squared error. This improvement is consistent across all seven effect sizes, with RMSE reductions ranging from 73\% (at $|\tau| = 0.30$) to 97.5\% (at $\tau = 0$). Notably, standard DML actually performs \emph{worse} than Naive OLS in terms of RMSE (0.317 vs.\ 0.168), illustrating that controlling for a leaky representation can be more harmful than not controlling for confounders at all.

Second, DICE-DML substantially reduces bias. Average absolute bias decreases from 0.115 for standard DML to 0.051 for DICE-DML, a 56\% reduction. While some residual bias remains, the improvement is substantial and consistent. For context, a bias of 0.051 in standardized units corresponds to a Cohen's $d$ of approximately 0.05, which is conventionally considered negligible.

Third, DICE-DML excels at identifying null effects. At the null effect point ($\tau = 0$), RMSE reduction reaches 97.5\%, indicating near-perfect recovery of the true zero effect. This performance is crucial for scientific inference: a method that cannot reliably identify the absence of an effect will generate false discoveries.

\begin{figure*}[t]
\centering
\includegraphics[width=0.75\textwidth]{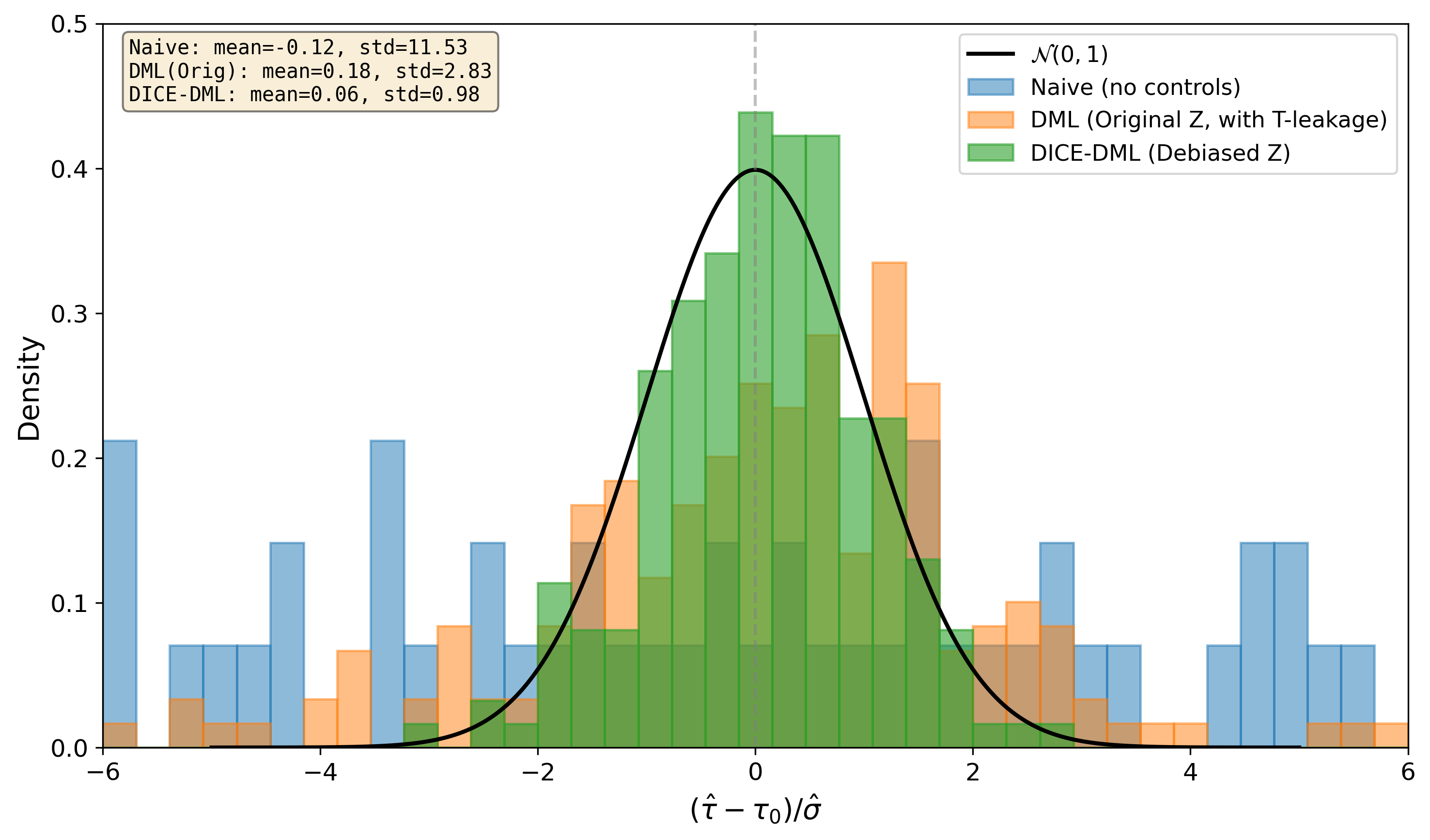}
\caption{Asymptotic Normality of Treatment Effect Estimators. Distribution of standardized t-statistics at $\tau = 0$ across 200 simulations. Naive estimation without controls (blue) produces highly dispersed estimates (std = 11.53). Standard DML with original ResNet features (orange) remains overdispersed (std = 2.83), indicating treatment leakage distorts variance. DICE-DML (green) closely tracks the theoretical $N(0,1)$ distribution (std $\approx$ 0.98), confirming valid inference and proper Type I error control.}
\Description{Histogram showing three distributions of t-statistics, with DICE-DML matching standard normal.}
\label{fig:normality}
\end{figure*}

A key theoretical property of DML estimators is asymptotic normality: under regularity conditions, the standardized estimation error $(\hat{\tau} - \tau_0)/\hat{\sigma}$ should converge to a standard normal distribution~\cite{chernozhukov2018dml}. Figure~\ref{fig:normality} visualizes this property by plotting the distribution of standardized t-statistics across 200 simulations at $\tau = 0$. Naive estimation without controls produces a highly dispersed distribution (std = 11.53), reflecting severe confounding. Standard DML with original ResNet features shows improvement but remains substantially overdispersed (std = 2.83), indicating that treatment leakage distorts the variance structure. DICE-DML closely tracks the theoretical $N(0,1)$ distribution with std $\approx 0.98$, confirming valid inference.

\subsection{Real-World Estimation}

We now apply DICE-DML to estimate the causal effect of skin tone on Instagram engagement in our real-world data, where the true effect is unknown. Table~\ref{tab:main} presents results comparing Naive OLS, standard DML with original ResNet features, and DICE-DML with our debiased representations.

\begin{table}[t]
\caption{Causal Effect Estimates for Skin Tone on Engagement}
\label{tab:main}
\begin{tabular}{lcccc}
\toprule
\textbf{Method} & \textbf{$\tau$ (likes)} & \textbf{95\% CI} & \textbf{$Y$ $R^2$} & \textbf{$T$ $R^2$} \\
\midrule
Naive OLS & $+74$ & --- & --- & --- \\
DML (Original) & $-1{,}455^{**}$ & $[-2494, -415]$ & $-0.003$ & 0.251 \\
\textbf{DML (DICE)} & $\mathbf{-522^\dagger}$ & $\mathbf{[-1070, +26]}$ & $\mathbf{0.626}$ & $\mathbf{0.149}$ \\
\bottomrule
\end{tabular}
\begin{flushleft}
\footnotesize \textit{Notes:} $^\dagger p < 0.10$, $^{**} p < 0.01$. $Y$ $R^2$: out-of-sample $R^2$ for outcome nuisance model. $T$ $R^2$: $R^2$ for propensity model.
\end{flushleft}
\end{table}

The most striking finding is the diagnostic contrast between standard DML and DICE-DML. Standard DML produces a negative outcome $R^2$ ($-0.003$), meaning the nuisance model for predicting engagement from the ResNet representation performs \emph{worse} than simply predicting the sample mean for every observation. This seemingly impossible result reveals a fundamental problem: the representation captures treatment-induced variation rather than genuine confounding variation. The nuisance model learns to predict this treatment-related ``noise,'' which does not generalize to held-out data. The high treatment $R^2$ (0.251) confirms the diagnosis: the ResNet representation strongly predicts treatment, indicating substantial leakage.

In contrast, DICE-DML achieves an outcome $R^2$ of 0.626, a qualitative transformation from model failure to successful confounding control. The representation now captures meaningful variation in engagement predictable from visual characteristics, which is exactly what a confounding control should do. Simultaneously, treatment $R^2$ falls to 0.149, confirming reduced (though not eliminated) leakage.

Naive OLS, which ignores confounding entirely, suggests essentially no raw association between skin tone and engagement ($+74$ likes, statistically indistinguishable from zero). Standard DML estimates a large, statistically significant negative effect ($-1,455$ likes; $p < 0.01$). However, given the diagnostic failure documented above, this estimate should not be trusted. DICE-DML estimates a smaller, marginally significant negative effect ($-522$ likes; $p = 0.062$). The 95\% confidence interval $[-1070, +26]$ includes zero, indicating that the evidence does not definitively rule out a null effect at conventional significance levels. The DICE estimate corrects approximately 64\% of the apparent bias in the standard DML estimate.

\subsection{Treatment Removal Visualization}

\begin{figure}[t]
\centering
\includegraphics[width=0.95\columnwidth]{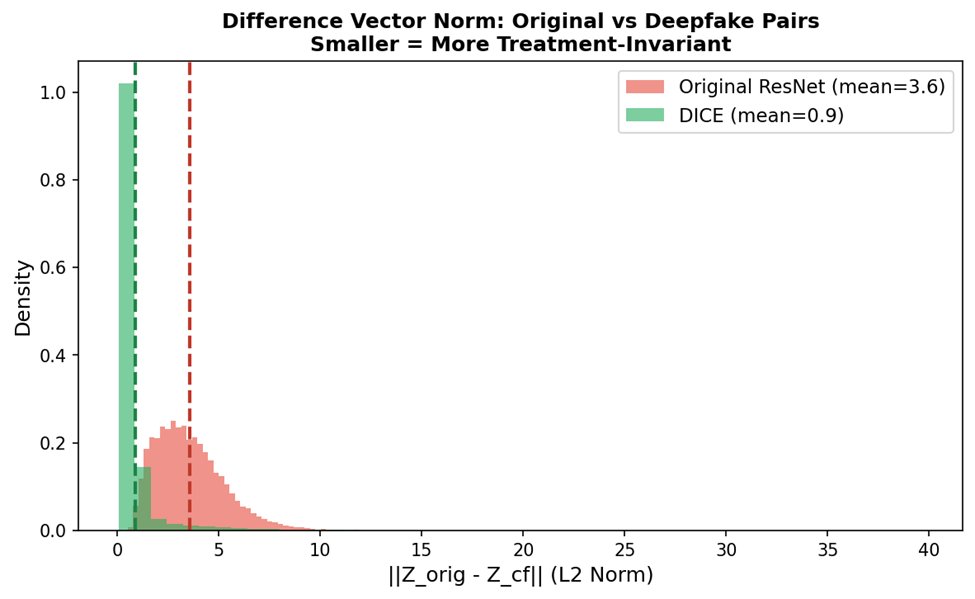}
\caption{Distribution of difference vector norms between original and deepfake image pairs. DICE-DML reduces the mean L2 distance from 3.6 to 0.9 (75\% reduction), indicating that the encoder learns to produce similar representations for image pairs that differ only in treatment (skin tone), while preserving other visual information. This compression does not reflect representation collapse, as the same embeddings achieve $Y$ $R^2 = 0.63$ for outcome prediction.}
\Description{Histogram comparing difference vector norm distributions for Original ResNet and DICE embeddings.}
\label{fig:diff_norm}
\end{figure}

To provide direct evidence that DICE-DML successfully removes treatment information from the representation, Figure~\ref{fig:diff_norm} visualizes the distribution of difference vector norms $\|Z_{\text{orig}} - Z_{\text{cf}}\|$ between original and deepfake image pairs. For the original ResNet embeddings, these norms follow a broad distribution centered around 3.6, indicating substantial variation between paired embeddings---variation that reflects encoded treatment information. After DICE processing, the distribution shifts dramatically leftward with mean 0.9, a 75\% reduction.

This compression is not merely geometric but has clear semantic interpretation: the encoder has achieved \emph{deepfake-pair invariance}, learning to produce nearly identical representations for image pairs that differ only in treatment (skin tone). Importantly, this compression does not reflect representation collapse: the same representations retain substantial outcome predictability ($Y$ $R^2 = 0.63$), confirming that only treatment-related variation is removed while outcome-relevant confounding information is preserved.

\subsection{Robustness}

A potential concern with DML is sensitivity to the choice of machine learning method for nuisance estimation. We assess robustness by re-estimating treatment effects using different nuisance model specifications: shallow Random Forest (50 trees, max depth 5), deep Random Forest (100 trees, max depth 10), Gradient Boosting Machine, and Ridge regression. Table~\ref{tab:robust} presents results for DICE-DML across these specifications.

\begin{table}[t]
\caption{Robustness to Nuisance Model Specification}
\label{tab:robust}
\begin{tabular}{lcccc}
\toprule
\textbf{Specification} & \textbf{$\tau$ (std)} & \textbf{SE} & \textbf{$\tau$ (likes)} & \textbf{$Y$ $R^2$} \\
\midrule
RF (shallow) & $-0.0106$ & 0.0097 & $-358$ & 0.716 \\
RF (deep) & $-0.0116$ & 0.0097 & $-391$ & 0.720 \\
GBM & $-0.0111$ & 0.0100 & $-375$ & 0.702 \\
Ridge & $-0.0103$ & 0.0154 & $-348$ & 0.341 \\
\bottomrule
\end{tabular}
\begin{flushleft}
\footnotesize \textit{Notes:} All specifications use DICE-encoded features. RF = Random Forest; GBM = Gradient Boosting Machine. $N = 50{,}000$.
\end{flushleft}
\end{table}

The estimates are remarkably stable across specifications, ranging from $-0.0103$ to $-0.0116$ in standardized units, corresponding to $-348$ to $-391$ likes. The standard deviation of point estimates across the four specifications is only 0.0005, negligible relative to the standard errors. This stability provides confidence that our findings do not depend on idiosyncratic properties of any particular machine learning method. All specifications achieve substantially positive outcome $R^2$, confirming that the DICE representation supports valid confounding control regardless of the downstream estimator. The Ridge regression specification, despite its lower $R^2$ due to linearity constraints, yields a point estimate virtually identical to the nonlinear methods, suggesting the core treatment effect signal is robust to functional form assumptions.

\subsection{Ablation Study}

To understand the contribution of each component in DICE-DML, we conduct an ablation study that systematically removes the two key mechanisms: DICE-Diff adversarial learning and orthogonal projection. Table~\ref{tab:ablation} demonstrates that the two components are not redundant but address orthogonal failure modes: adversarial learning preserves outcome-relevant variation, while orthogonal projection removes residual treatment signal.

\begin{table}[t]
\caption{Ablation Study: Component Contributions}
\label{tab:ablation}
\begin{tabular}{lccccc}
\toprule
\textbf{Configuration} & \textbf{RMSE} & \textbf{$Y$ $R^2$} & \textbf{$T$ $R^2$} & \textbf{$\|\Delta\|$} \\
\midrule
Baseline (encoder only) & 0.050 & 0.28 & 0.25 & 3.57 \\
Adversarial only & 0.042 & 0.90 & 0.14 & 3.57 \\
Projection only & 0.051 & $-$0.07 & 0.22 & 0.18 \\
\textbf{Full DICE-DML} & \textbf{0.043} & \textbf{0.63} & \textbf{0.15} & \textbf{0.87} \\
\bottomrule
\end{tabular}
\begin{flushleft}
\footnotesize \textit{Notes:} RMSE from simulation with known $\tau$. $\|\Delta\|$: mean L2 norm of $\|Z_{\text{orig}} - Z_{\text{cf}}\|$, measuring treatment information in representation. Lower $\|\Delta\|$ indicates better treatment removal. $N = 50{,}000$.
\end{flushleft}
\end{table}

The ablation reveals that the two components serve complementary functions. \textbf{Adversarial learning} primarily preserves outcome-predictive information: the ``Adversarial only'' configuration achieves the highest $Y$ $R^2$ (0.90) but fails to remove treatment information, as evidenced by the unchanged $\|\Delta\| = 3.57$ (identical to baseline). This high $\|\Delta\|$ indicates that the encoder still produces substantially different representations for original and deepfake image pairs, meaning treatment information remains encoded. Consequently, while the representation predicts outcomes well, it may do so partly by exploiting treatment signal---a form of leakage that compromises causal validity.

\textbf{Orthogonal projection} alone fails catastrophically: the ``Projection only'' configuration achieves low $\|\Delta\| = 0.18$, confirming geometric removal of the treatment axis, but $Y$ $R^2$ collapses to $-0.07$, worse than predicting the mean. This occurs because projection operates on the original ResNet features where treatment and confounding information are entangled; removing the treatment direction inevitably destroys outcome-relevant confounding signals. The elevated RMSE (0.051, worse than baseline) confirms that raw projection without learned disentanglement harms causal estimation.

The \textbf{full DICE-DML} combines both mechanisms to achieve the best bias-variance tradeoff. Adversarial learning first trains the encoder to disentangle treatment from confounding information into separable subspaces; orthogonal projection then cleanly removes the treatment component without collateral damage. The result: strong treatment removal ($\|\Delta\| = 0.87$, a 76\% reduction from baseline) while preserving substantial outcome predictability ($Y$ $R^2$ = 0.63). The simulation RMSE (0.043) confirms that only the combination achieves both valid confounding control and accurate effect estimation.

\section{Conclusion and Discussion}

This paper addresses the challenge of causal inference when the treatment is an attribute embedded within images, a setting increasingly relevant as visual content dominates digital marketing and questions of representation gain prominence. We identify \emph{visual treatment leakage}, the contamination of control representations by treatment information, as a fundamental problem for applying Double Machine Learning to image data. Standard vision encoders, trained to capture all visually salient information, inevitably embed treatment attributes alongside confounders. Using such representations as controls violates DML's identifying assumptions, producing biased estimates with characteristic diagnostic failure: negative outcome $R^2$ indicating that the nuisance model predicts noise rather than signal.

We develop DICE-DML (Deepfake-Informed Control Encoder for Double Machine Learning), which uses deepfake-generated image pairs as weak supervision for learning treatment-invariant representations. The key innovation, operating on difference vectors where background signals cancel, enables effective disentanglement of treatment from confounders without relying on spurious correlations. In simulation experiments with known ground-truth effects, DICE-DML reduces root mean squared error by 73--97\% compared to standard DML, with the strongest improvement (97.5\%) at the null effect point, demonstrating robust Type I error control. In our empirical application to Instagram influencer data, DICE-DML transforms the diagnostic profile from invalid (outcome $R^2 = -0.003$) to valid ($R^2 = 0.626$), representing a qualitative shift from model failure to successful confounding control.

Our substantive analysis of skin tone effects in Instagram advertising yields a marginally significant negative effect ($-522$ likes; $p = 0.062$), substantially smaller than the biased standard DML estimate ($-1,455$ likes). The 95\% confidence interval $[-1070, +26]$ includes zero, leaving uncertainty about whether any true penalty exists. For marketing practice, these findings suggest that diversity initiatives need not fear large engagement penalties from featuring diverse representation; the evidence does not support claims of substantial consumer backlash. At the same time, firms should not expect immediate engagement gains from diversity; the case for inclusive marketing may rest more on ethical and brand-building considerations than on short-term engagement metrics.

The methodological framework we develop extends beyond skin tone and Instagram. Visual treatment leakage arises whenever treatments are embedded within high-dimensional representations used as controls---sentiment in text, tone in audio, style in video. DICE-DML provides a template for addressing this challenge: use domain knowledge or generative models to construct paired examples isolating treatment variation, train representations that are invariant to this variation while preserving outcome-relevant information, and validate through diagnostic metrics before trusting causal estimates. The framework applies to any setting where approximate paired perturbations can be constructed to isolate a treatment dimension, even if such perturbations are imperfect and used only as weak supervision.

Several limitations warrant acknowledgment and suggest directions for future research. First, DICE-DML requires generating paired images, which may be infeasible for treatments that cannot be convincingly manipulated (e.g., complex scene characteristics). Developing alternative weak supervision strategies would extend applicability. Second, causal identification still relies on conditional independence given confounders; unobserved confounders (posting strategy, follower demographics, platform algorithms) remain a concern. Sensitivity analysis methods adapted to this setting would strengthen causal claims. Third, we study a single platform (Instagram), treatment (skin tone), and outcome (likes); generalization to other contexts requires further investigation. Fourth, extending DICE-DML to estimate heterogeneous effects across subgroups would provide richer insights for targeted marketing strategies.

\bibliographystyle{ACM-Reference-Format}
\bibliography{references}

\appendix

\section{Additional Implementation Details}

The variance regularization loss follows VICReg~\cite{bardes2022vicreg}:
\begin{equation}
\mathcal{L}_{\text{var}} = \frac{1}{d}\sum_{j=1}^{d}\max(0, \gamma - \text{std}(Z^{(j)}))
\end{equation}
where $\gamma = 1$ is the target standard deviation threshold and $d = 512$ is the representation dimension.

The outcome prediction loss uses the mean of paired representations:
\begin{equation}
\mathcal{L}_{Y} = \frac{1}{N}\sum_{i=1}^{N}(Y_i - h(\bar{Z}_i))^2
\end{equation}
where $h(\cdot)$ is a two-layer prediction head and $\bar{Z}_i = (Z_i + Z'_i)/2$.

The contrastive loss uses margin-based formulation:
\begin{equation}
\mathcal{L}_{\text{ctr}} = \frac{1}{N}\sum_{i=1}^{N}\max(0, m + \|Z_{\text{clean},i} - Z'_{\text{clean},i}\| - \|\bar{Z}_i - \bar{Z}_{\text{neg}(i)}\|)
\end{equation}
where $m = 0.5$ is the margin and $\text{neg}(i)$ indexes a random negative sample from the same batch.

\section{Ethical Considerations}

This research involves sensitive topics, skin tone measurement and synthetic image generation, that warrant explicit ethical discussion. First, our measurement of skin tone via ITA\textdegree{} is used solely for academic research to study potential disparities in platform engagement; the goal is to detect and quantify possible discrimination, not to enable it. Second, our use of image manipulation (``deepfakes'') serves exclusively as weak supervision for representation learning. We do not generate synthetic content for public distribution, impersonation, or deception; manipulated images exist only as intermediate training signals. Third, our data derive from publicly posted Instagram content collected by prior researchers following platform terms of service.

\section{Algorithm Details}

Algorithm~\ref{alg:dice} presents the complete DICE-DML training and estimation procedure.

\begin{algorithm}[H]
\caption{DICE-DML: Training and Estimation}
\label{alg:dice}
\begin{algorithmic}[1]
\REQUIRE Images $\{I_i\}_{i=1}^N$, treatments $\{T_i\}_{i=1}^N$, outcomes $\{Y_i\}_{i=1}^N$
\ENSURE Treatment effect estimate $\hat{\tau}$

\STATE \textbf{// Stage 1: Deepfake Generation}
\FOR{$i = 1$ to $N$}
    \STATE Generate $I'_i$ by altering skin tone of $I_i$
    \STATE Ensure $\text{ITA}(I'_i)$ crosses 28\textdegree{} threshold
\ENDFOR

\STATE \textbf{// Stage 2: Feature Extraction}
\STATE Extract ResNet features: $\tilde{Z}_i \leftarrow \text{ResNet}(I_i)$, $\tilde{Z}'_i \leftarrow \text{ResNet}(I'_i)$

\STATE \textbf{// Stage 3: DICE Encoder Training}
\STATE Initialize encoder $f_\theta$, discriminators $D_\phi$, $D_{\text{lin}}$, predictor $h_\psi$
\FOR{epoch $= 1$ to 30}
    \FOR{each mini-batch $\mathcal{B}$}
        \STATE $Z_i \leftarrow f_\theta(\tilde{Z}_i)$, $Z'_i \leftarrow f_\theta(\tilde{Z}'_i)$ for $i \in \mathcal{B}$
        \STATE $\Delta_i \leftarrow Z_i - Z'_i$ \COMMENT{Difference vectors}
        \STATE $\hat{\delta}_i \leftarrow \Delta_i / (\|\Delta_i\| + \epsilon)$ \COMMENT{Unit direction}
        \STATE \textbf{// Orthogonal Projection (applied during training)}
        \STATE $Z_{\text{clean},i} \leftarrow Z_i - \alpha(Z_i \cdot \hat{\delta}_i)\hat{\delta}_i$
        \STATE $Z'_{\text{clean},i} \leftarrow Z'_i - \alpha(Z'_i \cdot \hat{\delta}_i)\hat{\delta}_i$
        \STATE \textbf{// Compute losses}
        \STATE $\mathcal{L}_{\text{adv}} \leftarrow \text{BCE}(D_\phi(\Delta), \mathbf{1}) + \text{BCE}(D_{\text{lin}}(\Delta), \mathbf{1})$
        \STATE $\mathcal{L}_{\text{cons}} \leftarrow \|Z_{\text{clean}} - Z'_{\text{clean}}\|^2$
        \STATE $\mathcal{L}_{Y} \leftarrow \|Y - h_\psi((Z_{\text{clean}} + Z'_{\text{clean}})/2)\|^2$
        \STATE Update $\theta, \psi$ via gradient reversal on $\mathcal{L}_{\text{adv}}$
        \STATE Update $\phi$ to minimize discriminator loss
    \ENDFOR
\ENDFOR

\STATE \textbf{// Stage 4: Embedding Extraction}
\FOR{$i = 1$ to $N$}
    \STATE $Z_i \leftarrow f_\theta(\tilde{Z}_i)$, $Z'_i \leftarrow f_\theta(\tilde{Z}'_i)$
    \STATE $\hat{\delta}_i \leftarrow (Z_i - Z'_i) / (\|Z_i - Z'_i\| + \epsilon)$
    \STATE $Z_{\text{clean},i} \leftarrow Z_i - \alpha(Z_i \cdot \hat{\delta}_i)\hat{\delta}_i$
\ENDFOR

\STATE \textbf{// Stage 5: DML Estimation with Cross-Fitting}
\STATE Split data into $K=5$ folds
\FOR{$k = 1$ to $K$}
    \STATE Train $\hat{g}^{(-k)}$, $\hat{m}^{(-k)}$ on folds $\neq k$ using $Z_{\text{clean}}$
    \STATE Compute $\tilde{Y}_i = Y_i - \hat{g}^{(-k)}(Z_{\text{clean},i})$
    \STATE Compute $\tilde{T}_i = T_i - \hat{m}^{(-k)}(Z_{\text{clean},i})$
\ENDFOR
\STATE $\hat{\tau} \leftarrow [\sum_i \tilde{T}_i^2]^{-1} \sum_i \tilde{T}_i \tilde{Y}_i$

\RETURN $\hat{\tau}$
\end{algorithmic}
\end{algorithm}

\end{document}